%% file: acl2021.tex
\documentclass[11pt,a4paper]{article}
\usepackage[hyperref]{acl2021}
\usepackage{times}
\usepackage{latexsym}
\usepackage{tabularx}
\usepackage{graphicx}
\usepackage{multirow}
\usepackage{booktabs}
\usepackage{amsmath}
\usepackage{todonotes}
\usepackage{float}
\usepackage{enumitem}

\usepackage{booktabs,arydshln}

\makeatletter
\def\adl@drawiv#1#2#3{%
        \hskip.5\tabcolsep
        \xleaders#3{#2.5\@tempdimb #1{1}#2.5\@tempdimb}%
                #2\z@ plus1fil minus1fil\relax
        \hskip.5\tabcolsep}
\newcommand{\cdashlinelr}[1]{%
  \noalign{\vskip\aboverulesep
           \global\let\@dashdrawstore\adl@draw
           \global\let\adl@draw\adl@drawiv}
  \cdashline{#1}
  \noalign{\global\let\adl@draw\@dashdrawstore
           \vskip\belowrulesep}}
\makeatother

\usepackage{microtype}

\aclfinalcopy % Uncomment this line for the final submission
\def\aclpaperid{4237} %  Enter the acl Paper ID here

\definecolor{navyblue}{rgb}{0.0, 0.0, 0.5}
\definecolor{cadmiumorange}{rgb}{0.93, 0.53, 0.18}
\definecolor{darkspringgreen}{rgb}{0.09, 0.45, 0.27}
\definecolor{chamoisee}{rgb}{0.63, 0.47, 0.35}
\definecolor{ochre}{rgb}{0.8, 0.47, 0.13}

\title{Cross-document Coreference Resolution over Predicted Mentions}

\author{Arie Cattan\textsuperscript{1} \quad
    Alon Eirew\textsuperscript{1,2} \quad
    Gabriel Stanovsky\textsuperscript{3} \quad
    Mandar Joshi\textsuperscript{4} \quad
    Ido Dagan\textsuperscript{1}\\ \\
\textsuperscript{1}Computer Science Department, Bar Ilan University \\ 
\textsuperscript{2}Intel Labs, Israel \quad
\textsuperscript{3}The Hebrew University of Jerusalem 
 \\
\textsuperscript{4}Allen School of Computer Science \& Engineering, University of Washington, Seattle, WA
 \\
  {\tt  arie.cattan@gmail.com} \quad {\tt  alon.eirew@intel.com} \\ 
  {\tt gabis@cse.huji.ac.il} \quad
     {\tt mandar90@cs.washington.edu} \\ {\tt  dagan@cs.biu.ac.il}  
  }

\date{}

\begin{document}
\maketitle

\input{sections/abstract}
\input{sections/intro}

\input{sections/bg}

\input{sections/model}
\input{sections/experiments}
\input{sections/conclusion}
\input{sections/acknowledgement}
\input{sections/ethics}
\bibliographystyle{acl_natbib}
\bibliography{anthology, acl2021}

\input{sections/appendix}

\end{document}

%% file: sections/abstract.tex
\begin{abstract}
Coreference resolution has been mostly investigated within a single document scope, showing impressive progress in recent years based on end-to-end models. However, the more challenging task of cross-document (CD) coreference resolution remained relatively under-explored, with the few recent models applied only to gold mentions. 
Here, we introduce the first end-to-end model for CD coreference resolution from raw text, which extends the prominent model for within-document coreference to the CD setting. Our model achieves competitive results for event and entity coreference resolution on gold mentions. More importantly, we set first baseline results, on the standard ECB+ dataset, for CD coreference resolution over predicted mentions. Further, our model is simpler and more efficient than recent CD coreference resolution systems, while not using any external resources.\footnote{\url{https://github.com/ariecattan/coref}}
\end{abstract}

%% file: sections/intro.tex
\section{Introduction}

Cross-document (CD) coreference resolution consists of identifying textual mentions across multiple documents that refer to the same concept. For example, consider the following sentences from the ECB+ dataset~\citep{cybulska-vossen-2014-using}, where colors represent coreference clusters (for brevity, we omit some clusters): 

\vspace{-5pt}
\begin{enumerate}[leftmargin=*]
    \item \emph{\textbf{\textcolor{darkspringgreen}{Thieves}} \textbf{\textcolor{navyblue}{pulled off}} a two million euro jewellery heist in central Paris on Monday after \textbf{\textcolor{ochre}{smashing}} \textbf{\textcolor{darkspringgreen}{their}} car through the store's front window.}
    \item \emph{\textbf{\textcolor{darkspringgreen}{Four men}} \textbf{\textcolor{ochre}{drove}} a 4x4 through the front window of the store on Rue de Castiglione, before \textbf{\textcolor{navyblue}{making off}} with the jewellery and watches.}
\end{enumerate}
\vspace{-5pt}

Despite its importance for downstream tasks, CD coreference resolution has been lagging behind the impressive strides made in the scope of a single document~\citep{lee-etal-2017-end, joshi-etal-2019-bert, joshi-etal-2020-spanbert, wu-etal-2020-corefqa}. Further, state-of-the-art models exhibit several shortcomings, such as operating on gold mentions or relying on external resources such as SRL or a paraphrase dataset~\citep{shwartz-etal-2017-acquiring}, preventing them from being applied on realistic settings.

To address these limitations, we develop the first end-to-end CD coreference model building upon a prominent within-document (WD) coreference model~\cite{lee-etal-2017-end} which we extend with recent advances in transformer-based encoders.
We address the inherently non-linear nature of the CD setting by combining the WD coreference model with  agglomerative clustering that was shown useful in CD models. Our model achieves competitive results on ECB+ over gold mentions and sets baseline results over predicted mentions. Our model is also simpler and substantially more efficient than existing CD coreference systems. Taken together, our work seeks to bridge the gap between WD and CD coreference, driving further research of the latter in realistic settings.
% of CD coreference resolution.

%% file: sections/bg.tex
\section{Background}
\label{sec:bg}

\begin{figure*}[!ht]
\centering
\includegraphics[width=\textwidth]{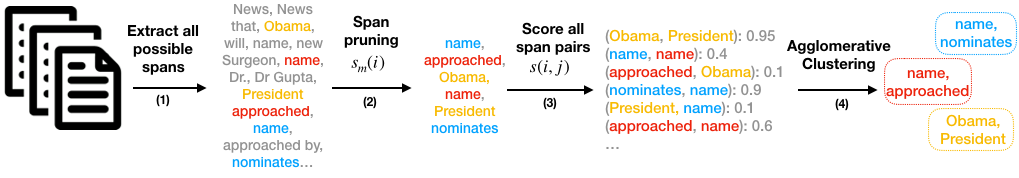}
\caption{A high-level diagram of our model for cross-document coreference resolution. (1) Extract and score all possible spans, (2) keep top spans according to $s_m(i)$, (3) score all pairs $s(i, j)$, and (4) cluster spans using agglomerative clustering.
\label{fig:diagram}
}\end{figure*}

\paragraph{Cross-document coreference}

Previous works on CD coreference resolution learn a pairwise scorer between mentions and use a clustering approach to form the coreference clusters~\citep{cybulska-vossen-2015-translating, yang-etal-2015-hierarchical, choubey-huang-2017-event, kenyon-dean-etal-2018-resolving, Bugert2020CrossDocumentEC}. \citet{barhom-etal-2019-revisiting} proposed to jointly learn entity and event coreference resolution, leveraging predicate-argument structures. Their model forms the coreference clusters incrementally, while alternating between event and entity coreference. 
% At each step, mention representations are updated according to the current clustering configuration, and cluster pairs with highest scores are merged. 
Based on this work,~\citet{meged-etal-2020-paraphrasing} improved results on event coreference by leveraging a paraphrase resource~(Chirps;~\citealp{shwartz-etal-2017-acquiring}) as distant supervision. 
Parallel to our work, recent approaches propose to fine-tune BERT on the pairwise coreference scorer~\citep{zeng-etal-2020-event}, where the state-of-the-art on ECB+ is achieved using a cross-document language model (CDLM) on pairs of full documents~\citep{Caciularu2021CrossDocumentLM}.
Instead of applying BERT for all mentions pairs which is quadratically costly, our work separately encodes each (predicted) mention.

All above models suffer from several drawbacks. First, they use only gold mentions and treat entities and events separately.\footnote{Few works~\citep{yang-etal-2015-hierarchical, choubey-huang-2017-event} do use predicted mentions, by considering the intersection of predicted and gold mentions for evaluation, and thus not penalizing models for false positive mention identification. Moreover, they used a different version of ECB+ with known annotation errors, as noted in~\citep{barhom-etal-2019-revisiting}.} Second, pairwise scores are recomputed after each merging step, which is resource and time consuming. Finally, they rely on additional resources, such as semantic role labeling, a within-document coreference resolver, and a paraphrase resource, which limits the applicability of these models in new domains and languages. In contrast, we use no such external resources.

\paragraph{Within-document coreference}

The \emph{e2e-coref} WD coreference model~\cite{lee-etal-2017-end} learns for each span $i$ a distribution over its antecedents.

Considering all possible spans as potential mentions, the scoring function $s(i, j)$ between span $i$ and $j$, where $j$ appears before $i$, has three components: the two mention scores $s_m(\cdot)$ of spans $i$ and $j$, and a pairwise antecedent score $s_a(i, j)$ for span $j$ being an antecedent of span $i$. 

Each span is represented with the concatenation of four vectors: the output representations of the span boundary (first and last) tokens ($x_\text{FIRST(i)}, x_\text{LAST(i)}$), an attention-weighted sum of token representations $\hat{x}_i$, and a feature vector $\phi(i)$.
These span representations ($g_i$) are first fed into a mention scorer $s_m(\cdot)$ to filter the $\lambda T$ (where $T$ is the number of tokens) spans with the highest scores.
Then, the model learns for each of these spans to optimize the marginal log-likelihood of its correct antecedents. The full description of the model is described below:
\begin{align*} 
\label{eq:span_repr}
g_{i} &= [x_\text{FIRST(i)}, x_\text{LAST(i)}, \hat{x}_i, \phi(i) ] \\
s_{m}(i) &= \mathnormal{ \text{FFNN}_m(g_i)} \\
s_{a}(i, j) &=  \text{FFNN}_a([g_i, g_j, g_i \circ g_j]) \\
s(i, j) &= s_m(i) + s_m(j) + s_a(i, j) 
\end{align*}

%% file: sections/model.tex
\section{Model}
\label{sec:model}

The overall structure of our model is shown in Figure~\ref{fig:diagram}.
The major obstacle in applying the \emph{e2e-coref} model directly in the CD setting is its reliance on \emph{textual ordering} -- it forms coreference chains by linking each mention to an antecedent span appearing before it in the document. This linear clustering method cannot be used in the multiple-document setting since there is no inherent ordering between the documents. Additionally, ECB+ (the main benchmark for CD coreference resolution) is relatively small compared to OntoNotes~\citep{pradhan-etal-2012-conll}, making it hard to jointly optimize mention detection and coreference decision. These challenges have implications in all stages of model development, as elaborated below.

\paragraph{Pre-training} To address the small scale of the dataset, we pre-train the mention scorer $s_m(\cdot)$ on the gold mention spans, as ECB+ includes singleton annotation. This enables generating good candidate spans from the first epoch, and as we show in Section~\ref{subsec:ablations}, it substantially improves performance.

\paragraph{Training} 
Instead of comparing a mention only to its previous spans in the text, our pairwise scorer $s_a(i, j)$ compares a mention to all other spans across all the documents.
The positive instances for training consist of all the pairs of highest scoring mention spans that belong to the same coreference cluster, while the negative examples are sampled (20x the number of positive pairs) from all other pairs. The overall score is then optimized using the binary cross-entropy loss as follows:

\begin{align*}
    L = -\frac{1}{|N|} \sum_{x, z \in N} {y \cdot log(s(x, z))}
\end{align*}
where N corresponds to the set of mention-pairs $(x, z)$, and $y \in \{0, 1\}$ to a pair label. Full implementation details are described in Appendix~\ref{app:model}.
Notice that the mention scorer $s_m(\cdot)$ is further trained in order to generate better candidates at each training step. When training and evaluating the model in experiments over gold mentions, we ignore the span mention scores, $s_m(\cdot)$, and the gold mention representations are directly fed into the pairwise scorer $s_a(i, j)$.

\paragraph{Inference}
At inference time, we score all spans; prune spans with lowest scores; score the pairs; and finally form the coreference clusters using an agglomerative clustering (using average-linking method) over these pairwise scores, following common practices in CD coreference resolution~\cite{yang-etal-2015-hierarchical, choubey-huang-2017-event, kenyon-dean-etal-2018-resolving, barhom-etal-2019-revisiting}. 
Since the affinity scores $s(i, j)$ are also computed for mention pairs in different documents, the agglomerative clustering can effectively find cross-document coreference clusters.

%% file: sections/experiments.tex
\section{Experiments}

\subsection{Experimental setup}
\label{subsec:exp}

Following most recent work, we conduct our experiments ECB+~\cite{cybulska-vossen-2014-using}, which is the largest dataset that includes both WD and CD coreference annotation (see Appendix~\ref{app:dataset}). We use the document clustering of~\citet{barhom-etal-2019-revisiting} for pre-processing and apply our coreference model separately on each predicted document cluster. 

Following~\citet{barhom-etal-2019-revisiting}, we present the model's performance on both event and entity coreference resolution. In addition, inspired by~\citet{lee-etal-2012-joint}, we train our model to perform event and entity coreference jointly, which we term ``ALL". This represents a useful scenario when we are interested in finding all the coreference links in a set of documents, without having to distinguish event and entity mentions. Addressing CD coreference with ALL is challenging because (1) the search space is larger than when treating separately event and entity coreference and (2) models need to make subtle distinctions between event and entity mentions that are lexically similar but do not corefer. For example, the entity \emph{voters} do not corefer with the event \emph{voted}.

We apply RoBERTa\textsubscript{\emph{LARGE}}~\cite{liu2019roberta} to encode the documents. Long documents are split into non-overlapping segments of up to 512 word-piece tokens and are encoded independently \cite{joshi-etal-2019-bert}. Due to memory constraints, we freeze output representations from RoBERTa instead of fine-tuning all parameters. For all experiments, we use a single GeForce GTX 1080 Ti 12GB GPU. The training takes 2.5 hours for the most expensive setting (ALL on predicted mentions), while inference over the test set takes 11 minutes.

\subsection{Results}
\label{subsec:results}
\input{tables/results}
\input{tables/wd_results}

Table~\ref{tab:results} presents the combined within- and cross-document results of our model, in comparison to previous work on ECB+. We report the results using the standard evaluation metrics MUC, B\textsuperscript{3}, CEAF, and the average F1 of these metrics, called CoNLL F1 (main evaluation).

When evaluated on gold mentions, our model achieves competitive results for event (81 F1) and entity (73.1) coreference. In addition, we set baseline results where the model does not distinguish between event and entity mentions at inference time (denoted as the ALL setting). The overall performance on ECB+ obtained using two separate models for event and entity is negligibly higher (+0.6 F1) than our single ALL model.

Our model is the first to enable end-to-end CD coreference on \emph{raw} text (predicted mentions). As expected, the performance is lower than that using gold mentions (e.g 26.6 F1 drop in event coreference), indicating the large room for improvement over predicted mentions. It should be noted that beyond mention detection errors, two additional factors contribute to the performance drop when moving to predicted mentions. First, while WD coreference systems typically disregard singletons (mentions appearing only once) when evaluating on \emph{raw} text, CD coreference models do consider singletons when evaluating on \emph{gold} mentions on ECB+. We observe that this difference affects the evaluation, explaining about 10\% absolute points out of the aforementioned drop of 26.6. The effect of singletons on coreference evaluation is further explored in~\citep{cattan2021eval}. Second, entities are annotated in ECB+ only if they participate in event, making participant detection an additional challenge. This explains the more important performance drop in entity and ALL.

Table~\ref{tab:wd_results} presents the CoNLL F1 results of within- and cross-document coreference resolution for both gold and predicted mentions on ECB+. For all settings, results are higher in within-document coreference resolution, showing the need in addressing typical challenges of CD coreference resolution.
\input{tables/without_dc}

Table~\ref{tab:results_without_dc} shows the results of our model without document clustering. Here, the performance drop and error reduction are substantially larger for event coreference (-6.2/12\%) than entity coreference (-1.3/2\%) and ALL (-2/3.5\%). This difference is probably due to the structure of ECB+ which poses a lexical ambiguity challenge for events, while the document clustering step reconstructs almost perfectly the original subtopics, as shown in~\citep{barhom-etal-2019-revisiting}.

Further, the higher results on event coreference do not mean that the task is inherently easier than entity coreference. In fact, when ignoring singletons in the evaluation, as done on OntoNotes, the performance of event coreference is lower than entity coreference (62.1 versus 65.3 CoNLL F1)~\citep{cattan2021eval}. This happens because event singletons are more common compared to entity singletons (30\% vs. 17\%), as shown in Appendix~\ref{app:dataset}.

Finally, our model is more efficient in both training and inference since the documents are encoded using just one pass of RoBERTa, and the pairwise scores are computed only once using a simple MLP. For comparison, previous models compute pairwise scores at each iteration~\citep{barhom-etal-2019-revisiting, meged-etal-2020-paraphrasing}, or apply a BERT-model to every mention pairs with their sentence~\citep{zeng-etal-2020-event} or full document~\citep{Caciularu2021CrossDocumentLM}.\footnote{For a rough estimation, our model runs for 2 minutes while \citet{barhom-etal-2019-revisiting}'s model runs for 37 minutes on similar hardware. }

\subsection{Ablations}
\label{subsec:ablations}

To show the importance of each component of our model, we ablate several parts and compute F1 scores on the development set of the ECB+ event dataset. The results are presented in Table~\ref{tab:ablations} using predicted mentions without document clustering.

\input{tables/ablations}

Skipping the \emph{pre-training} of the mention scorer results in a 3.2 F1 points drop in performance. Indeed, the relatively small training data in the ECB+ dataset (see Appendix~\ref{app:dataset}) might be not sufficient when using only end-to-end optimization, and pre-training of the mention scorer helps generate good candidate spans from the first epoch.

To analyze the effect of the dynamic pruning, we froze the mention scorer during the pairwise training, and kept the same candidate spans along the training. The significant performance drop (4 F1) reveals that the mention scorer inherently incorporates coreference signal.

Finally, using all negative pairs for training leads to a performance drop of 1.4 points and significantly increases the training time.

\subsection{Qualitative Analysis}

We sampled topics from the development set and manually analyzed the errors of the ALL configuration.
The most common errors were due to an over-reliance on lexical similarity. For example, the event ``\emph{Maurice Cheeks was \textbf{fired}}" was wrongly predicted to be coreferent with a similar, but different event, ``\emph{the Sixers \textbf{fired} Jim O'Brien}", probably because of related context, as both coached the Philadelphia 76ers. On the other hand, the model sometimes struggles to merge mentions that are lexically different but semantically similar (e.g ``\emph{Jim O'Brien \textbf{was shown the door}}", ``\emph{Philadelphia \textbf{fire} coach Jim O'Brien}"). The model also seems to struggle with temporal reasoning, in part due to missing information. For example, news articles from different days have different relative reference to time, while the publication date of the articles is not always available. As a result, the model missed linking ``\emph{Today}" in one document to ``\emph{Saturday}" in another document.

%% file: tables/results.tex
\begin{table*}[!h]
    \centering
    \resizebox{\textwidth}{!}{
    \begin{tabular}{lllccclccclccclc}
    \toprule
    \phantom{defgrnbkjb} & \phantom{defgrnbkjb} && \multicolumn{3}{c}{MUC} && \multicolumn{3}{c}{B\textsuperscript{3}} && \multicolumn{3}{c}{$CEAFe$} &&
    CoNLL\\
    \cmidrule{4-6} \cmidrule{8-10} \cmidrule{12-14} \cmidrule{16-16}
    &&& R & P & $F_1$ && R & P & $F_1$ && R &P & $F_1$ && $F_1$  \\
    \midrule
    \multirow{6}{*}{Event} & \citet{barhom-etal-2019-revisiting} && 77.6 & 84.5& 80.9 && 76.1 & 85.1 & 80.3 && 81.0 & 73.8 & 77.3 && 79.5 \\
    & \citet{meged-etal-2020-paraphrasing} && 78.8 & 84.7 & 81.6 && 75.9 & 85.9 & 80.6 && 81.1 & 74.8 & 77.8 && 80.0 \\
    & Our model \textendash{}  Gold && 85.1 & 81.9 & 83.5 && 82.1 & 82.7 & 82.4 && 75.2 & 78.9 & 77.0 &&  \emph{81.0} \\
    & \citet{zeng-etal-2020-event}$^{*}$ && 85.6 & 89.3 & 87.5 && 77.6 & 89.7 & 83.2 && 84.5 & 80.1 & 82.3 && 84.6 \\
    & \citet{Caciularu2021CrossDocumentLM}$^{*}$ && 87.1  & 89.2 & 88.1 && 84.9 & 87.9 & 86.4 && 83.3 & 81.2 & 82.2  && \textbf{85.6} \\ 
    \cdashlinelr{2-16}
    & Our model \textendash{} Predicted && 66.6 & 65.3 & 65.9 && 56.4 & 50.0 & 53.0 && 47.8 & 41.3 & 44.3 && \emph{\textbf{{54.4}}} \\
    \midrule
    \multirow{4}{*}{Entity} & \citet{barhom-etal-2019-revisiting} && 78.6 & 80.9 & 79.7 && 65.5 & 76.4 & 70.5 && 65.4 & 61.3 & 63.3 && 71.2 \\
    & Our model \textendash{}  Gold && 85.7 & 81.7 & 83.6 && 70.7 & 74.8 & 72.7 && 59.3 & 67.4 & 63.1 && \emph{73.1} \\
    & \citet{Caciularu2021CrossDocumentLM}$^{*}$  && 88.1 & 91.8 & 89.9 && 82.5 & 81.7 & 82.1 && 81.2 & 72.9 & 76.8  && \textbf{82.9} \\
    \cdashlinelr{2-16}
    & Our model \textendash{}  Predicted && 43.5 & 53.1 & 47.9 && 25.0 & 38.9 & 30.4 && 31.4 & 26.5 & 28.8 && \emph{\textbf{{35.7}}} \\
    \midrule
    \multirow{2}{*}{ALL} & Our model \textendash{} Gold && 84.2 & 81.6 & 82.9 && 76.8 & 77.5 & 77.1 && 68.4 & 72.4 & 70.3 && \emph{\textbf{76.7}} \\
    \cdashlinelr{2-16}
    & Our model \textendash{}  Predicted && 49.7 & 58.5 & 53.7 && 33.2 & 46.5 & 38.7 && 40.4 & 35.2 & 37.6 && \emph{\textbf{{43.4}}} \\
    \bottomrule
    \end{tabular}}
    \caption{Combined within- and cross-document results on the ECB+ test set, for event, entity and the unified task, that we term ALL. 
    Our results (in italics) are close to state-of-the-art (in bold) for event and entity over gold mentions, while they set a new benchmark result over predicted mentions and for the ALL setting. The run-time complexity in \citep{zeng-etal-2020-event, Caciularu2021CrossDocumentLM} is substantially more expensive because they apply BERT and CDLM for every mention-pair with their corresponding context (sentence and full document).}
    \label{tab:results}
\end{table*}

%% file: tables/wd_results.tex
\begin{table}
    \centering
    \resizebox{0.4\textwidth}{!}{
    \begin{tabular}{lccccc}
        \toprule
        \phantom{aaaaaaaaaa} & \multicolumn{2}{c}{Gold} && \multicolumn{2}{c}{Predicted}  \\
        & WD & CD && WD & CD \\
         \midrule
        Event & \textbf{86.6} & 81.0 && \textbf{59.6} & 54.4 \\
        Entity & \textbf{81.2} & 73.1 && \textbf{39.7} & 35.7 \\
        ALL & \textbf{83.9} & 76.7 && \textbf{46.3} & 43.4 \\
        \bottomrule
    \end{tabular}}
    \caption{Results (CoNLL F1) of our model, on within-document (WD) vs. cross-document (CD), using gold and predicted mentions. For all settings, results on WD are higher, indicating the need in addressing typical challenges of CD coreference resolution.}
    \label{tab:wd_results}
\end{table}

%% file: tables/without_dc.tex
\begin{table}
    \centering
    \resizebox{0.48\textwidth}{!}{
    \begin{tabular}{lccccc}
        \toprule
        \phantom{aaaaaaaaaa} & \textbf{Gold} & $\Delta$ & \phantom{aa} &  \textbf{Predicted} & $\Delta$ \\
         \midrule
        Event & 76.0 & \textbf{--5.0} && 48.2 & \textbf{--6.2} \\
        Entity & 70.9 & --2.2 && 34.4 & --1.3 \\
        ALL & 74.1 & --2.6 && 41.4 & --2.0 \\
        \bottomrule
    \end{tabular}}
    \caption{CoNLL F1 results of our model without document clustering , using gold and predicted mentions. }
    \label{tab:results_without_dc}
\end{table}

%% file: tables/ablations.tex
\begin{table}[!t]
    \newcommand{\colindent}{\;}
    \centering
    \resizebox{0.45\textwidth}{!}{
    \begin{tabular}{lcc}
    \toprule
    \phantom{fwidsvhckzxjvchndfzxvvgdaczfc} & F1 & $\Delta$\\
    \midrule
    Our model  & 58.1 & \\
    \colindent $-$ pre-train of mention scorer & 54.9 & $-$3.2 \\
    \colindent $-$ dynamic pruning & 54.1 & $-$4.0 \\
    % \colindent $-$ RoBERTa & 54.0 & $-$4.1\\
    \colindent $-$ negative sampling & 56.7 & $-$1.4 \\
    \bottomrule
    \end{tabular}}
    \caption{Ablation results (CoNLL F1) of our model on the development set of ECB+ event coreference.     }
    \label{tab:ablations}
\end{table}

%% file: sections/conclusion.tex
\section{Conclusion and Discussion}

We developed the first end-to-end baseline for CD coreference resolution over predicted mentions. Our simple and efficient model achieve competitive results over gold mentions on both event and entity coreference, while setting baseline results for future models over predicted mentions. 

Nonetheless, we note a few limitations of our model that could be addressed in future work. First, following most recent work on cross-document coreference resolution (§\ref{sec:bg}), our model requires $\mathcal{O}(n^2$) pairwise comparisons to form the coreference cluster. While our model is substantially more efficient than previous work (§\ref{subsec:results}), applying it on a large-scale dataset would involve a scalability challenge. Future work may address the scalability issue by using recent approaches for hierarchical clustering on massive datasets~\citep{Monath2019ScalableHC, Monath2021ScalableHC}. Another appealing approach consists of splitting the corpus into subsets of documents, constructing initial coreference clusters (in parallel) on the subsets, then merging meta-clusters from the different sets. 
We note though that it is currently impossible to test such solutions for more extensive scalability, pointing to a need in collecting larger-scale datasets for cross-document coreference.
Second, to improve overall performance over predicted mentions, future work may incorporate, explicitly or implicitly, semantic role labeling signals in order to identify event participants for entity prediction, as well as for better event structure matching. Further, dedicated components may be developed for mention detection and coreference linking, which may be jointly optimized.

%% file: sections/acknowledgement.tex
\section*{Acknowledgments}
We thank the anonymous reviewers for their helpful comments. The work described herein was supported in part by grants from Intel Labs, Facebook, the Israel Science Foundation grant 1951/17, the Israeli Ministry of Science and Technology, the German Research Foundation through the German-Israeli Project Cooperation (DIP, grant DA 1600/1-1), and from the Allen Institute for AI.

%% file: sections/ethics.tex
\section*{Ethical Considerations}

\paragraph{Dataset} As mentioned in the paper, we use the ECB+ dataset, available at \url{http://www.newsreader-project.eu/results/data/the-ecb-corpus/}. 

\paragraph{Model} Our cross-document coreference model does not contain any intentional biasing or ethical issues. As mentioned in the paper (§\ref{subsec:exp}), we conduct our experiments on a single 12GB GPU, and both training and inference times are relatively low. 

%% file: sections/appendix.tex
\clearpage
\appendix

\section{Appendix}

\subsection{Implementation Details}
\label{app:model}

Our model includes 14M parameters and is implemented in PyTorch \cite{NEURIPS2019_9015}, using HuggingFace's library~\cite{wolf-etal-2020-transformers} and the Adam optimizer \cite{kingma2014adam}. 
The layers of the models are initialized with Xavier Glorot method \cite{glorot2010understanding}.  
We manually tuned the standard hyperparameters, presented in Table~\ref{tab:shared} on the event coreference task and keep them unchanged for entity and ALL settings.
Table~\ref{tab:specific} shows specific parameters, such as the maximum span width, the pruning coefficient $\lambda$ and the stop criterion $\tau$ for the agglomerative clustering, that we tuned separately for each setting to maximize the CoNLL F1 score on its corresponding development set.

\input{tables/hyperparam}

\subsection{Dataset}
\label{app:dataset}

ECB+\footnote{\url{http://www.newsreader-project.eu/results/data/the-ecb-corpus/}} is an extended version of the EventCorefBank (ECB) \citep{bejan-harabagiu-2010-unsupervised} and EECB~\citep{lee-etal-2012-joint}, whose statistics are shown in Table~\ref{tab:ecb_stat}. The dataset is composed of 43 \emph{topics}, where each topic corresponds to a famous news event (e.g Someone checked into rehab). In order to introduce some complexity and to limit the use of lexical features, each topic is constituted by a collection of texts describing two different event instances of the same event type, called \emph{subtopic}. For example, the first topic corresponding to the event \emph{``Someone checked into rehab"} is composed of event mention of the event \emph{``Tara Reid checked into rehab"} and \emph{``Lindsay Lohan checked into rehab"} which are obviously annotated into different coreference cluster. Documents in ECB+ are in English. Since ECB+ is an \emph{event-centric} dataset, entities are annotated only if they participate in events. In this dataset, event and entity coreference clusters are denoted separately.

\input{tables/ecb_stats}

%% file: tables/hyperparam.tex
\begin{table}[!h]
    \centering
    \begin{tabular}{lll}
    \toprule
    Hyperparameter   &\phantom{abckjbn}  &  Value\\
    \midrule
    Batch size && 32 \\
    Dropout && 0.3 \\
    Learning rate && 0.0001\\
    Optimizer && Adam \\
    Hidden layer && 1024 \\
    \bottomrule
    \end{tabular}
    \caption{Shared hyperparameters across the different models. }
    \label{tab:shared}
\end{table}

\begin{table}[!h]
    \centering
    \begin{tabular}{lllllll}
    \toprule
    && Max span width && $\lambda$ && $\tau$ \\
    \midrule
    Event  &&  10 && 0.25 && 0.75 \\
    Entity  &&  15 && 0.35 && 0.75 \\
    ALL  &&  15 && 0.4 && 0.75 \\
    \bottomrule
    \end{tabular}
    \caption{Specific hyperparameters for each mention type; $\lambda$ is the pruning coefficient and $\tau$ is the threshold for the agglomerative clustering. }
    \label{tab:specific}
\end{table}

%% file: tables/ecb_stats.tex
\begin{table}[!h]
    \centering
    \resizebox{0.48\textwidth}{!}{
    \begin{tabular}{@{}llll@{}}
    \toprule
    & \textbf{Train} & \textbf{Validation} & \textbf{Test} \\
    \midrule
    \# Topics & 25 & 8 & 10 \\
    \# Documents & 594 & 196 & 206 \\
    \# Mentions & 3808/4758 & 1245/1476 & 1780/2055 \\
    \# Singletons & 1116/814 & 280/205 & 632/412 \\
    \# Clusters & 1527/1286 & 409/330 & 805/608 \\
    \bottomrule
    \end{tabular}}
    \caption{ECB+ statistics. The slash numbers for \# Mentions, \# Singletons and \# Clusters represent event/entity statistics. As recommended by the authors in the release note, we follow the split of \citet{cybulska-vossen-2015-translating} that use a curated subset of the dataset.}
    \label{tab:ecb_stat}
\end{table}